\documentclass[10pt,letterpaper]{article}
\usepackage[top=0.85in,left=2.75in,footskip=0.75in]{geometry}

\usepackage{changepage}

\usepackage[utf8]{inputenc}

\usepackage{textcomp,marvosym}

\usepackage{fixltx2e}

\usepackage{amsmath,amssymb}

\usepackage{cite}

\usepackage{nameref,hyperref}

\usepackage[right]{lineno}

\usepackage{microtype}
\DisableLigatures[f]{encoding = *, family = * }

\usepackage{rotating}

\raggedright
\usepackage[aboveskip=1pt,labelfont=bf,labelsep=period,justification=raggedright,singlelinecheck=off]{caption}

\makeatletter
\renewcommand{\@biblabel}[1]{\quad#1.}
\makeatother

\date{}

\usepackage{float}

\usepackage{multirow}
\usepackage{rotating}

\usepackage{subfigure}
\usepackage{color}

\definecolor{myblue}{rgb}{.0,.0,.8}
\newcommand{\added}[1]{\textcolor{black}{#1}}

\newcommand{\quantile}[1]{\text{quantile}#1}

\begin{document}

\title{Automatic Emphysema Detection using Weakly Labeled HRCT Lung Images}

\author{Isabel Pino Pe{\~n}a*\textsuperscript{1\Yinyang},
Veronika Cheplygina*\textsuperscript{2,3\Yinyang},
Sofia Paschaloudi\textsuperscript{4\ddag},
Morten Vuust\textsuperscript{4\ddag},
Jesper Carl\textsuperscript{5\ddag},
Ulla M{\o}ller Weinreich\textsuperscript{5,6\ddag},
Lasse Riis {\O}stergaard\textsuperscript{1\ddag},
Marleen de Bruijne\textsuperscript{3,7\ddag}
\\
\bigskip
\textbf{1} Department of Health Science and Technology, Aalborg University, Aalborg, Denmark
\\
\textbf{2} Department of Biomedical Engineering, Eindhoven University of Technology, Eindhoven, The Netherlands
\\
\textbf{3} Biomedical Imaging Group Rotterdam, Erasmus Medical Center, Rotterdam, The Netherlands
\\
\textbf{4} Department of Diagnostic Imaging, Vendsyssel Hospital, Fredrikshavn, Denmark
\\
\textbf{5} Department of Clinical Medicine, Aalborg University Hospital, Aalborg, Denmark
\\
\textbf{6} Department of Pulmonary Medicine, Aalborg University Hospital, Aalborg, Denmark
\\
\textbf{7} Department of Computer Science, University of Copenhagen, Copenhagen, Denmark
}

\vspace*{0.35in}

\begin{flushleft}
{\Large
\textbf\newline{Automatic emphysema detection using weakly labeled HRCT lung images}
}\newline
\\
Isabel Pino Pe{\~n}a*\textsuperscript{1\Yinyang},
Veronika Cheplygina*\textsuperscript{2,3\Yinyang},
Sofia Paschaloudi\textsuperscript{4\ddag},
Morten Vuust\textsuperscript{4\ddag},
Jesper Carl\textsuperscript{5\ddag},
Ulla M{\o}ller Weinreich\textsuperscript{5,6\ddag},
Lasse Riis {\O}stergaard\textsuperscript{1\ddag},
Marleen de Bruijne\textsuperscript{3,7\ddag}
\\
\bigskip
\textbf{1} Department of Health Science and Technology, Aalborg University, Aalborg, Denmark
\\
\textbf{2} Department of Biomedical Engineering, Eindhoven University of Technology, Eindhoven, The Netherlands
\\
\textbf{3} Biomedical Imaging Group Rotterdam, Erasmus Medical Center, Rotterdam, The Netherlands
\\
\textbf{4} Department of Diagnostic Imaging, Vendsyssel Hospital, Fredrikshavn, Denmark
\\
\textbf{5} Department of Clinical Medicine, Aalborg University Hospital, Aalborg, Denmark
\\
\textbf{6} Department of Pulmonary Medicine, Aalborg University Hospital, Aalborg, Denmark
\\
\textbf{7} Department of Computer Science, University of Copenhagen, Copenhagen, Denmark
\\
\bigskip

\Yinyang These authors contributed equally to this work.

\ddag These authors also contributed equally to this work.

* Corresponding authors:  ipino@hst.aau.dk (IPP), v.cheplygina@tue.nl (VC)

\end{flushleft}










\begin{abstract}
Purpose: A method for automatically quantifying emphysema regions using High-Resolution Computed Tomography (HRCT) scans of patients with chronic obstructive pulmonary disease (COPD) that does not require manually annotated scans for training is presented.

Methods: HRCT scans of controls and of COPD patients with diverse disease severity are acquired at two different centers. Textural features from co-occurrence matrices and Gaussian filter banks are used to characterize the lung parenchyma in the scans. Two robust versions of multiple instance learning (MIL) classifiers that can handle weakly labeled data, miSVM and MILES, are investigated. Weak labels give information relative to the emphysema without indicating the location of the lesions. The classifiers are trained with the weak labels extracted from the forced expiratory volume in one minute (FEV$_1$) and diffusing capacity of the lungs for carbon monoxide (DLCO). At test time, the classifiers output a patient label indicating overall COPD diagnosis and local labels indicating the presence of emphysema. The classifier performance is compared with manual annotations made by two radiologists, a classical density based method, and pulmonary function tests (PFTs).

Results: The miSVM classifier performed better than MILES on both patient and emphysema classification. The classifier has a stronger correlation with PFT than the density based method, the percentage of emphysema in the intersection of annotations from both radiologists, and the percentage of emphysema annotated by one of the radiologists. The correlation between the classifier and the PFT is only outperformed by the second radiologist.

Conclusions: The presented method uses MIL classifiers to automatically identify emphysema regions in HRCT scans. Furthermore, this approach has been demonstrated to correlate better with DLCO than a classical density based method or a radiologist, which is known to be affected in emphysema. Therefore, it is relevant to facilitate assessment of emphysema and to reduce inter-observer variability.

\end{abstract}


\section{Introduction}

Chronic obstructive pulmonary disease (COPD) is the most important respiratory disease worldwide and one of the most important causes of death in high and middle-income countries~\cite{Mathers06, WHO16}. COPD is described as a progressive and irreversible airflow limitation. Emphysema is one of the most common disease manifestations that causes this limitation due to the destruction of alveolar walls and loss of elasticity~\cite{GOLD2015}. Emphysema can be identified visually in computed tomography (CT) scans as low attenuation areas (LAA). However, to enable the detection of lesions smaller than 5 mm, thin slice reconstructions, such as high-resolution computed tomography (HRCT) scans, are preferred.

The automatic identification and quantification of emphysema provides objectivity and more reliability to the clinical routine in the assessment of COPD. Currently, emphysema is assessed visually, which is time consuming, subjective and suffers from inter- and intra-observer variability~\cite{Ginsburg12}. Over the years, the most used methods for automatically quantifying emphysema have been density based~\cite{Matsuoka10, Lynch09, Nakano00}. These methods use a threshold based on percentile density or LAA, generally lower than -950 Hounsfield units (HU). However, these methods are very dependent on, among others, the inspiration level, scanner reconstruction kernel, exposure dose and scanners. Therefore, there is no consensus on the best threshold for quantifying emphysema~\cite{nishimura98, mets12}. Other quantification methods have been reported based on texture features, which collect information about the spatial relationship of the intensity values in the scan~\cite{soren10, nagao07, Yao11}. Machine learning methods based on texture analysis extract information to learn normal and abnormal lung tissues, which facilitates the recognition of disease patterns and can therefore lead to a more reliable diagnosis~\cite{Bagci12}. In general, machine learning methods use supervised classifiers that require annotated regions of interest (ROIs) or labeled patches based on manual annotations of emphysema performed by clinical experts~\cite{prasad07, Uppaluri99, park08, kim09}. Manual annotations are even more time consuming than visual assessment of emphysema and also suffer from inter-observer variation~\cite{chabat03}.

Learning from weak labels, which assign a label to the entire image, is proposed in the literature as the less time-consuming alternative to the manual annotation of patches, and it is being increasingly used in different medical image analysis applications~\cite{cheplygina14,melendez2014novel,kandemir2014computer}. Weak labels are easier to acquire than manual annotations because they can be obtained from basic quantification methods or complimentary data of the patient, such as pulmonary function tests (PFTs) or bio-markers. Classifiers which learn from weak labels are referred to as multiple instance learning (MIL) classifiers. All MIL classifiers can learn to label entire scans. For example, S{\o}rensen et al.~\cite{soren12}
and Cheplygina et al.~\cite{cheplygina14} used spirometry results, which is the most common PFT to clinically assess COPD, to assign labels to scans from the Danish Lung Cancer Screening Trial, and trained different types of MIL classifiers to detect COPD in previously unseen scans from the same trial. However, a subset of MIL classifiers can also learn to classify individual patches, thus identifying regions with signs of COPD, including emphysema. Neither~\cite{soren12}
nor~\cite{cheplygina14} evaluated MIL classifiers for this purpose. For example, more than half of the classifiers studied in~\cite{cheplygina14} including the best performing classifier, could not provide individual patch labels.

In contrast with previous studies, this study aims to automatically identify emphysema regions in patients with COPD using HRCT scans without local annotations. Different texture-based methods and MIL classifiers are investigated. Furthermore, in this study, more robust versions of two MIL classifiers are proposed. The results from the classifiers are evaluated with manual annotations made by two radiologists.

\section{Materials and Methods}
\begin{figure*}[ht]
		\centering
		\includegraphics[width=1\textwidth]{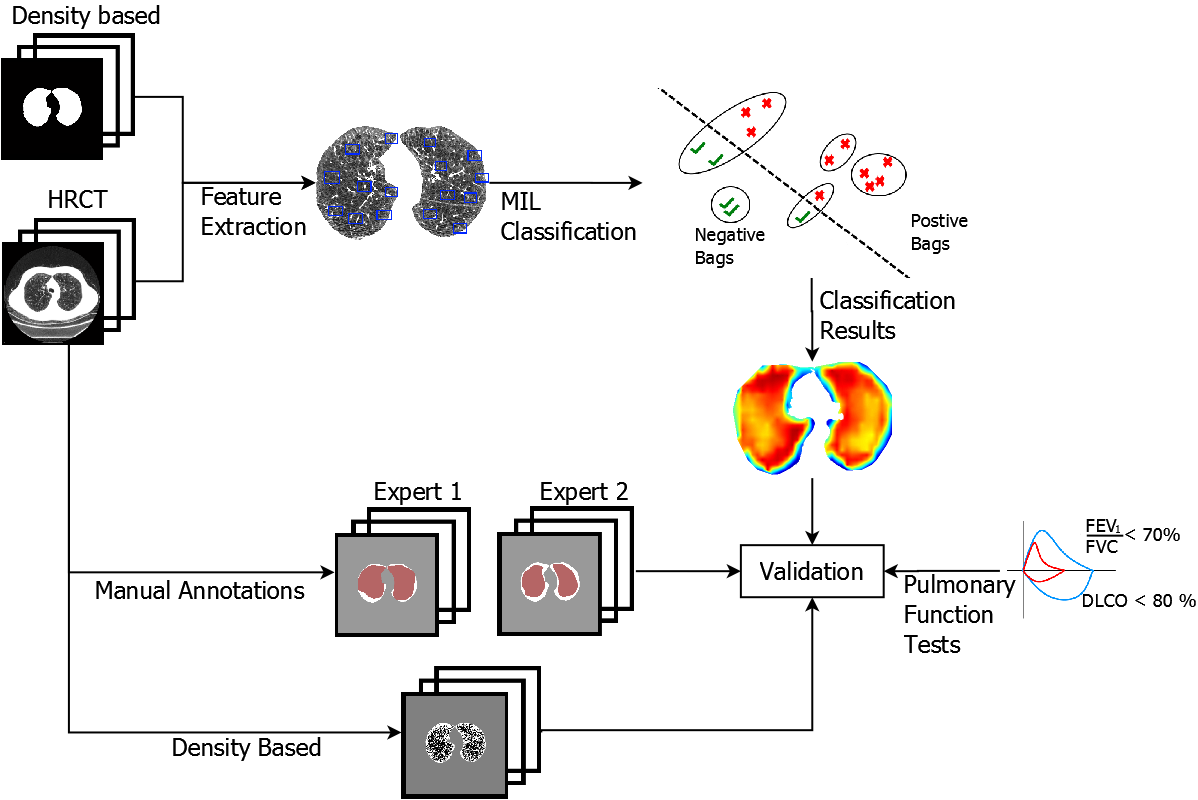}
	{\caption{Summary of the methodology. Texture features are extracted from the lung parenchyma. Two different MIL classifiers are trained and are tested on previously unseen scans. The results are evaluated against manual annotations performed by two radiologists, a density based analysis, and pulmonary function tests.}\label{fig:methodOverview}}
\end{figure*}

This study focuses on automatically distinguishing emphysema without using manual annotations to train the classifiers. For this purpose, different types of texture features are extracted to characterize emphysema, and two variations of MIL classifiers are investigated. Fig \ref{fig:methodOverview}
presents an overview of the method used.


\subsection{Features}

Two different types of texture features are computed: features from co-occurrence matrices and Gaussian derivative features. The co-occurrence matrix algorithm is used in 3D, and it aims to capture the spatial dependence of gray-level intensities through multiple slices. The co-occurrence of voxel pairs is evaluated in 13 directions and at five different distances. After obtaining the co-occurrence matrices, the spatial dependencies of gray-level values are described by 12 Haralick textural features: energy, entropy, correlation, contrast, homogeneity, variance, sum mean, inverse difference moment, inertia, cluster shade, cluster tendency and max probability~\cite{albregtsen08}.

Gaussian derivative features aim to capture the presence of structures such as edges and blobs. Each image is first convolved (using normalized convolution) with a Gaussian function: $G(\mathbf{v},\sigma) = \frac{1}{(((2\pi)^{1/2}\sigma)^3)} \exp{(- \frac{(||\mathbf{v}||^2_2)}{(2\sigma^2)})}$, where $\sigma$ represents the standard deviation of the Gaussian, or the scale at which the texture is examined, and $\mathbf{v}=[x,y,z]^T$ is a voxel. Similar to~\cite{soren12}, eight filters are computed: smoothed image, gradient magnitude, Laplacian of Gaussian, three eigenvalues of the Hessian, Gaussian curvature, and eigen magnitude. The filters are computed at four different scales: 0.6mm, 1.2mm, 2.4mm, and 4.8mm. The filtered outputs are summarized using histograms with ten bins, where the bin sizes are determined by adaptive binning~\cite{ojala1996comparative} on an independent dataset~\cite{soren12} prior to this study.

\subsection{Classifiers}

MIL is originally a binary classification problem, although multi-class extensions also exist~\cite{zhou2006multi}. MIL classifiers are trained on labeled \emph{bags} $\{(B_i, y_i) | i=1,...N\}$, where $i$ indicates the $i$-th out of total $N$ subjects, and $y_i$ is the label ($y_i = +1$ for COPD, or $y_i = -1$ for non-COPD) of the $i$-th subject. The bags are also referred to as positive or negative. Each bag $B_i  = \{\mathbf{x}_{ij}| j=1,...,n_i\} \subset \mathbb{R}^d$, is a set of $n_i$ texture feature vectors or \emph{instances}, where $\mathbf{x}_{ij}$ describes the $j$-th patch of the $i$-th subject.

In this study, the bags represent the entire scan of an individual subject, whereas the instances are randomly selected 3D patches from inside the lungs. The bags are related to the weak labels extracted from the pulmonary function tests, and the instance labels classify the lung parenchyma into emphysematous or healthy lung tissue. Typically, MIL classifiers assume that there are instance labels $y_{ij}$ that relate to the bag labels as follows: a bag is positive if and only if it contains at least one positive or \emph{concept} instance: $y_i = \max_j y_{ij}$. Thus, a bag is classified as positive if at least one instance contains emphysematous tissue. In this study a less strict assumption is used, as described in Section~\ref{sec:avoiding}.

There are two main strategies that MIL classifiers follow~\cite{amores2013multiple,cheplygina14}. The instance-level strategy is to use the bag labels to infer an instance classifier. To classify a previously unseen test bag, such classifiers classify its instances and then combine the instance labels into a bag label. The bag-level strategy is to represent the bags by some global characteristics and use supervised classifiers to classify the bags directly. Inferring the instance labels from the bag labels is not always possible in this case. In this study, the posterior probability that a classifier outputs for a bag is denoted as $f(B_i)$ and the posterior probability that a classifier outputs for an instance as $f(\mathbf{x}_{ij})$.

Two popular and effective MIL classifiers are miSVM~\cite{andrews2002support} and MILES~\cite{chen2006miles}. The instance-level miSVM classifier extends the traditional support vector machine (SVM) by searching for an instance classifier that separates the instances as well as possible but such that the $\max_j{\{y_{ij}\}} = y_i$ condition still holds. In other words, the most positive (according to the classifier) instance in each bag is positive if the bag is positive and negative if the bag is negative. Similar to a supervised SVM, the miSVM can operate on polynomial or radial basis kernels between instances. A regularization parameter $C$ controls the trade-off between the margin, i.e. how well the instances are separated, and how many training bags are incorrectly classified with this margin. For a test bag, its instances are classified, and the most positive instance determines the label of the bag.

MILES is a bag-level approach that is able to infer instance labels. It assumes that positive and negative bags contain discriminative prototype instances. It represents each bag by a feature vector $\mathbf{s}_i$ that contains its similarities to all instances in the training set, where the similarity is defined as  $s(B_i,\mathbf{x}) = \max_j{ k(\mathbf{x}_{ij}, \mathbf{x})}$, in which $k$ is a similarity function between instances, i.e., a polynomial or radial basis kernel. The maximum operator implies that the bag's similarity to an instance is high if it contains a single similar instance. The MILES classifier then selects discriminative similarities, which correspond to discriminative prototype instances. A regularization parameter $C$ controls the trade-off between how many bags are incorrectly classified and how many discriminative prototypes are selected. For a test bag, the similarity to the discriminative prototypes determine whether the bag is positive (if it has instances sufficiently similar prototypes from positive bags).

\subsection{Avoiding false positives}\label{sec:avoiding}

Because a single positive instance is sufficient to classify whether a bag is positive, miSVM and MILES may suffer from false positives. In this study, more robust formulations of miSVM and MILES are proposed, which we refer to as miSVM-Q and MILES-Q, which use the quantile rather than the maximum operator to define the label of the bag. In miSVM-Q,  $\max_j{\{y_{ij}\}} = y_i$ is replaced  by $\quantile_j{(   \{y_{ij}\}, q)} = y_i$, where $q$ is the desired quantile. For example, if $q=0.5$, half of the instances must be positive for a bag to be positive.

In MILES-Q, the similarity function to $s(B_i,\mathbf{x}) = \quantile_j{( \{k(\mathbf{x}_{ij}, \mathbf{x}) \}, q)}$ is adapted. This means that the bag must contain more similar instances to the prototype $\mathbf{x}$ to be considered similar to it. For both miSVM-Q and MILES-Q, these adaptations mean that healthy subjects can still be considered healthy if they have a few emphysematous patches.

Furthermore, this study proposes an additional measure to evaluate a MIL classifier $f$, which is called Separability or $S$:
    \begin{equation}
        S =  \frac{1}{\sum_{y_i=+1} n_i}  \sum_{y_i = + 1} f(\mathbf{x}_{ij}) -  \frac{1}{\sum_{y_i=-1} n_i} \sum_{y_i = - 1} f(\mathbf{x}_{ij}).
        \label{eq:separability}
    \end{equation}

In other words, the Separability describes the difference between the average posterior probabilities of instances in true positive training bags, and the average posterior probabilities of instances in true negative training bags. The intuition behind this is that true positive bags should have a larger proportion of positive instances, and therefore the average instance posterior probability should be higher than in negative bags. This allows reasoning about the classifier's performance on instance-level, without having access to instance labels.

Consider two classifiers $f_1$ and $f_2$ which classify a positive and a negative bag, each with two instances. For the positive bag, $f_1$ outputs posteriors 0.51 and 0.49, and $f_2$ outputs posteriors 0.9 and 0.1. For the negative bag, $f_1$ outputs 0.49 and 0.49, while $f_2$ outputs 0.1 and 0.1. While both $f_1$ and $f_2$ correctly classify the bags, $f_2$ would be the preferred classifier on instance-level. The Separability of $f_1$ and $f_2$ are respectively 0.01 and 0.4, which reflects our preference for $f_2$. Since Separability is a difference of two averages, each of which is between 0 and 1, Separability could theoretically range between -1 and 1. In our experiments we observed that most values fall between 0.05 and 0.75.

\section{Experimental}

\subsection{Data} \label{sec:experimental_data}

Two datasets are used in this study. Both datasets are named after the hospital where the HRCT scans were performed: Frederikshavn (Fre) and Aalborg (Aal). For both datasets, volumetric (helical) HRCT scans and pulmonary function tests (PFTs) are performed.

The PFTs are spirometry and diffusing capacity of the lungs for carbon monoxide (DLCO) and are acquired for each patient. PFTs and HRCT scans are performed with the patients in a steady state, i.e., no exacerbation within six weeks prior to the test, and HRCT scans are acquired with the patients in the supine position and with breath held. No contrast agents are used.


Volumetric (helical) HRCT scans from both datasets are acquired with the patients in supine position and with breath hold. No contrast agents are used. In the Frederikshavn dataset, the scans are performed on a Siemens SOMATOM Definition Flash CT scanner with scan parameters as follows: 0.6 slice thickness, 95 mAs, 120 kvp, rotation time 0.5 seconds, CTDIvol 7.96 mGy, pitch 1.2 with a image voxel resolution of 0.58$\times$0.588$\times$0.6 mm. In the Aalborg dataset, the scans are achieved using a GE 160 Discovery CT750 HD scanner with with scan parameters as follows: 0.625 slice thickness, 120 kvp, rotation time 0.5 seconds, CTDIvol 5.12 mGy, pitch 0.984, automatically calculated mA by GE´s Smart mA system (max 300 mA and Noise Index 40) and image voxel resolution of 0.788$\times$0.788$\times$0.6 mm.

Note that the dose is higher for the Siemens scan than for the GE scan, \added{however the scans were visually inspected by the two radiologists who performed the clinical validation and they concluded that the visual quality of the scans is similar.}






Table \ref{tab:Demographic} presents the clinical characteristics of both datasets. The Fre dataset contains COPD subjects and non-COPD subjects. The non-COPD subjects are referred from the out-patient clinic to have a HRCT scan due to different respiratory problems. Aal contains only subjects with COPD.

\begin{table*}[ht]
\centering
\caption{Clinical characteristics of subjects belonging to both datasets. GOLD stratification reflects the classification of the COPD patients according to the \textit{GOLD combined risk stratification assessment~\cite{GOLD2015}.}}
\resizebox{\textwidth}{!}{\begin{tabular}{l l l l lll llll l l}

Dataset &  &  Gender & Age & \multicolumn{3}{ c }{Smoking} & \multicolumn{4}{ c }{GOLD Stratification} & FEV$_1$ (\%) & DLCO (\%)\\
&  & (M/F)  &  & current & former & never & A & B & C & D & & \\  \hline
\multirow{2}{*}{Fre} & COPD & 7/1 & 66 [48-77] & 1 & 7 & 0 & 1 & 3 & 3 & 1 & 58 [36-91] & 55 [32-90] \\
 & non-COPD & 3/5 & 56 [25-73] & 1 & 2 & 5 & & & & & 96 [63-137] & 74 [62-83] \\
Aal & & 34/38 & 66 [32-83] & 23 & 48 & 1 & 24 & 12 & 18 & 18 & 62 [18-111] & 55 [15-108] \\
\end{tabular}}
\label{tab:Demographic}
\end{table*}

\subsection{Experimental Setup}
Two sets of experiments are performed, which differ in the ways that the positive and negative bags are defined. There are two ways to define the bag labels: by thresholding the COPD stratification (A-D = positive, otherwise = negative), and by thresholding the DLCO predicted value ($<$60\% = low (positive), $>$60\% = high (negative)). The value of $60\%$, 
rather than the traditional value of $80\%$ 
is chosen due to the small sample of subjects with DLCO $>$80\%. 
Thus, patients with lower DLCO (which could be indicative of COPD, but does not define the COPD diagnosis) are included in the high DLCO class.

To train and evaluate the classifiers, 50 patches with a size of $41\times41\times41$ pixels are randomly extracted from each HRCT scan. Previous studies~\cite{cheplygina14,soren09} have demonstrated that 50 patches is sufficient to classify an entire scan. The patches are selected inside the lung parenchyma using the lung masks.

In each patch, textural features extracted from co-occurrence matrices and Gaussian filter banks are computed. A total of 780 features are computed from co-occurrence matrices, and 320 features are obtained from Gaussian filter banks. High-dimensional feature representations are chosen because previous studies with MIL classifiers and similar feature representations~\cite{soren12,cheplygina14} showed good results in terms of bag-level performance. We will also briefly discuss our experiences with lower-dimensional features in the Results section.


\subsubsection{Cross-validation}

For each set of experiments, a 4-fold stratified cross-validation is performed, such that each fold contains a similar distribution of subjects. 
Thus, Fre and Aal datasets are combined and each fold contains non-COPD and COPD subjects with varying degrees of COPD severity, as well as subjects with low and high DLCO values.

The 4-fold cross validation uses 3 folds for training and the fourth fold for evaluation. During training on the 3 folds, an internal 3-fold cross-validation is done to optimize the parameters. These parameters, which are selected only using the 3 folds of the training set, are then used to train a classifier on all the 3 training set folds. The classifier is evaluated on the fourth fold. This is repeated 4 times, so each of the folds is used once for evaluation.

MILES-Q and miSVM-Q classifiers are investigated, and the best parameters for each classifier are selected using the training set. The parameter ranges for both classifiers are as follows: polynomial kernel $p \in \{1, 2\}$, radial basis kernel $rbf \in \{8, 10, 12, 14, 16, 20\}$, regularization parameter $C \in \{0.001, 0.003, 0.01, 0.03, 0.1, 0.3, 1\}$, and quantile parameter $q \in \{0.25, 0.5, 0.75, 0.9, 1\}$.

\subsection{Evaluation}

\subsubsection{Classifier evaluation}\label{sec:classifier_evaluation}

During the 4-fold cross-validation, the best combination of parameters for each classifier is extracted on the three training folds. For the test results, the bag AUC (area under the receiver-operating curve) and Separability (Eq.~\ref{eq:separability}) are examined. 
The bag AUC expresses the ability of the classifier to rank a randomly drawn positive HRCT scan higher than randomly drawn negative HRCT scan (i.e., assign a higher posterior probability to a positive scan). Therefore, AUC is not sensitive to class imbalance: if a classifier assigns all cases to the majority class, the accuracy would be high, but the AUC would be equal to 0.5. 
The Separability reflects the classifier's ability to distinguish patches with signs of COPD and healthy lung tissue, without having access to such labels. Performance is considered good when the bag AUC is as high as possible and the Separability is as large as possible.

\subsubsection{Clinical validation}
In the clinical validation, the set of features and the classifier with the best performance in terms of bag AUC and Separability on the training sets are chosen for each of the test folds. The classifier is tested on 10 slices per HRCT scan. This number of slices is chosen to keep manual annotations by the radiologists feasible. The slices are spaced 25 slices apart in each HRCT scan, avoiding the slices belonging to the top and bottom parts of the lungs. In selected slices, the classifier classifies every 10th voxel in both directions in the slice, that is, inside the lung mask for that slice.

Two radiologists, expert 1 and expert 2, with 40 and ten years of experience, respectively, working with HRCT scans on a daily basis annotated all emphysema lesions in the same ten slices per scan in which the classifier is tested. The manual annotations are performed using OsiriX imaging software (www.osirix-viewer.com) using a medical display (BARCO E-2621). The annotation process is blinded. Thus, the experts do not know the outlines of the other expert or the classification results. The amount of emphysema annotated by each expert, in percentage, is computed, as is the percentage of emphysema on which both experts agreed. 



For local emphysema detection, the default threshold of 0.5 is used to transform the posterior probabilities into emphysema or healthy category labels.

Spearman correlation analysis is performed, in which the emphysema percentages of the classifier are compared with the manual annotations, results from spirometry and DLCO, and a simple method based on the threshold of LAA. The threshold is set to -950 HU, which has been demonstrated to be an acceptable threshold for density based emphysema quantification~\cite{Wang13}. A comparison of correlations from independent samples using the Fisher r-to-z transformation is computed to assess the significance of the differences between the results from the Spearman correlation.

The inter-observer variability between experts is also investigated using the Dice similarity coefficient (DSC), which is a measurement of similarity. The values of DSC range from 0 if there is no agreement to 1 if there is a perfect match.

\section{Results}
\subsection{Classifier performance} \label{sec:classifier_performance}
As shown in Table \ref{tab:clasfresults}, miSVM-Q has higher performance than MILES-Q on both bag AUC and Separability. For miSVM-Q, Gaussian features provide larger Separability and generally better bag classification than co-occurrence features. The combination of co-occurrence matrices and Gaussian features does not improve the results obtained with these features alone. In general, both classifiers can better distinguish obstructions given by low FEV$_1$ (ClassCOPD) than by low DLCO (ClassDLCO).

Additionally, lower-dimensional feature combinations are briefly investigated, such as orientation-invariant co-occurrence features (60 features) and using only a histogram of intensities (10 features); however, the results were worse than those obtained using the full co-occurrence or Gaussian features.

This suggests that, similar to earlier studies~\cite{soren12,cheplygina14} the relatively high feature dimensionality is not detrimental to these classifiers.

\begin{table*}[ht]
\begin{adjustwidth}{-2.25in}{0in} 
\centering
\caption{miSVM-Q and MILES-Q results using COPD (ClassCOPD) and DLCO (ClassDLCO) labels. S: separability ($\times$100); AUC: bag AUC ($\times$100).}
\begin{tabular}{l |  rr| rr || rr | rr}

&\multicolumn{4}{ c }{miSVM-Q} & \multicolumn{4}{ c }{MILES-Q}\\

&\multicolumn{2}{ c }{DLCO} &  \multicolumn{2}{ c }{COPD}  & \multicolumn{2}{ c }{DLCO} &  \multicolumn{2}{ c }{COPD}  \\

Feature & AUC  & S & AUC & S & AUC & S & AUC & S  \\
\hline


Cooc & 70.9 $\pm$ 6.3 & 4.1 & \textbf{ 100.0 $\pm$ 0.0} & 44.2 & 53.0 $\pm$ 10.3 & 0.7 & \textbf{ 93.8 $\pm$ 6.2 }& 17.1 \\
Gauss & \textbf{ 81.6 $\pm$ 10.2 }& \textbf{ 21.7} & \textbf{ 100.0 $\pm$ 0.0 }& \textbf{ 61.1} & \textbf{ 69.1 $\pm$ 6.6 }& \textbf{ 7.0} & 89.4 $\pm$ 6.4 & \textbf{ 27.8} \\
Both & 59.5 $\pm$ 5.4 & 2.9 & 95.0 $\pm$ 3.5 & 19.1 & 50.8 $\pm$ 11.3 & -0.1 & 78.8 $\pm$ 18.2 & 17.5 \\

\end{tabular}

\label{tab:clasfresults}
\end{adjustwidth}
\end{table*}

\begin{figure}[!ht]
		\centering
		\includegraphics[width=1\columnwidth]{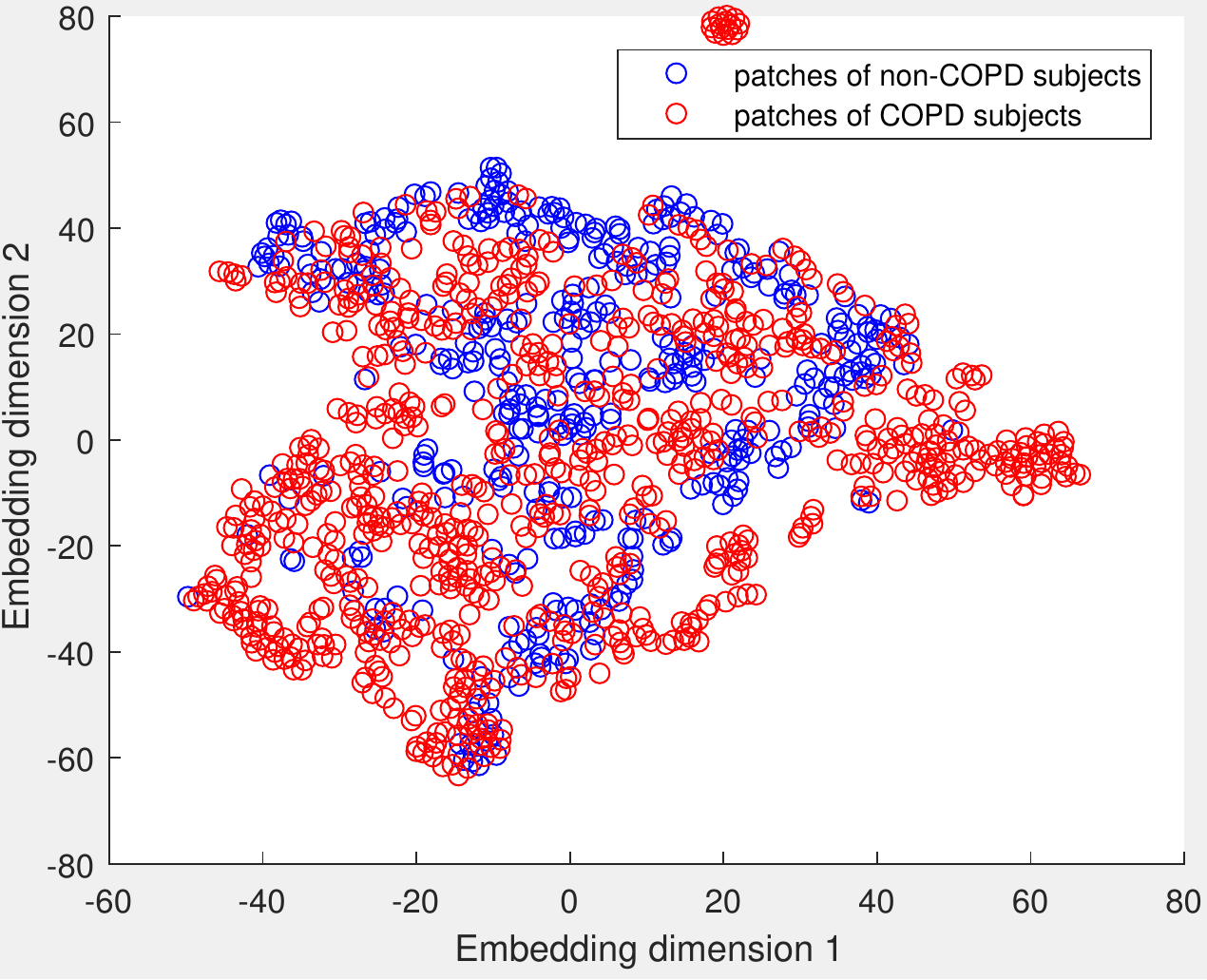}
	\caption{2D visualization of patches from COPD and non-COPD subjects using the Gaussian feature representation. The patches have different distributions, which helps the MIL classifier to classify a subject correctly.}
	\label{fig:tsne}
\end{figure}

To support the classifier evaluation, we additionally visualize one of the high-dimensional feature spaces (Gaussian texture features), which consists of patches from COPD subjects and non-COPD subjects in Fig~\ref{fig:tsne}.
Note that, due to the MIL representation, there is no straightforward way to cluster the subjects themselves. This dimensionality reduction and visualization is performed using t-SNE~\cite{maaten2008visualizing}, a popular method which aims to represent the local structure of the patches in the high-dimensional space as well as possible, without using the labels of the patches. The labels are only added to the plot for visualization. \added{Note that the embedding is the same after rotation, there are therefore no meaningful names that can be assigned to the dimensions, similar to principal component analysis.}


We use all the patches from non-COPD subjects, and a random sample (due to class imbalance) of patches from the COPD subjects, for an uncluttered visualization. The visualization shows a large overlapping area with patches that both types of subjects have. However, on the periphery (bottom left, and the right side of the plot) there are regions which only contain patches from COPD subjects. A subject having patches in these areas could therefore be classified as belonging to the COPD class.

\subsection{Association with PFTs} 
Based on the classifier evaluation results, miSVM-Q and Gaussian features are selected for the clinical validation. The parameters of miSVM-Q for the different test folds are selected during cross-validation as before. 

Table \ref{tab:SpearmanClass} presents the percentage of emphysema detected by the different methods and their correlation with DLCO and FEV$_1$. The correlations are considered significant at the 0.05 level. Moreover, analysis using the Fisher r-to-z transformation is computed which shows that there is not a significant difference between the correlation coefficients from the Spearman analysis.

\begin{table*}[ht]
\centering
\caption{Spearman correlation results with data from pulmonary tests. ClassCOPD: results from classifier with COPD label; ClassDLCO: results from classifier with DLCO label; Thr LAA: Threshold scan based on low attenuation areas; Agree Exp: area of agreement between the manual annotations of both experts; rho: correlation coefficient.}
\resizebox{\textwidth}{!}{\begin{tabular}{l l | r r r | r r r}

 &  & ClassCOPD & ClassDLCO & Thr LAA & Agree Exp & Expert1 & Expert2 \\ \hline

 \multirow{2}{*}{DLCO val} & rho & -0.477 & -0.571 & -0.513 & -0.478 & -0.472 & \textbf{-0.596} \\
  & $p$ Value & $<$0.0001 & $<$0.0001 & $<$0.0001 & $<$0.0001 & $<$0.0001 & $<$.0001 \\
\multirow{2}{*}{FEV$_1$} & rho & -0.283 & -0.383 & \textbf{-0.461} & -0.298 & -0.316 & -0.314 \\
 & $p$ Value & 0.016 & $<$0.0001 & $<$0.0001 & 0.011 & 0.007 & 0.007 \\

\end{tabular}}
\label{tab:SpearmanClass}
\end{table*}

\subsection{Association with manual annotations} \label{ManAnn}
\begin{figure*}[ht]
		\centering
		\includegraphics[width=0.32\textwidth]{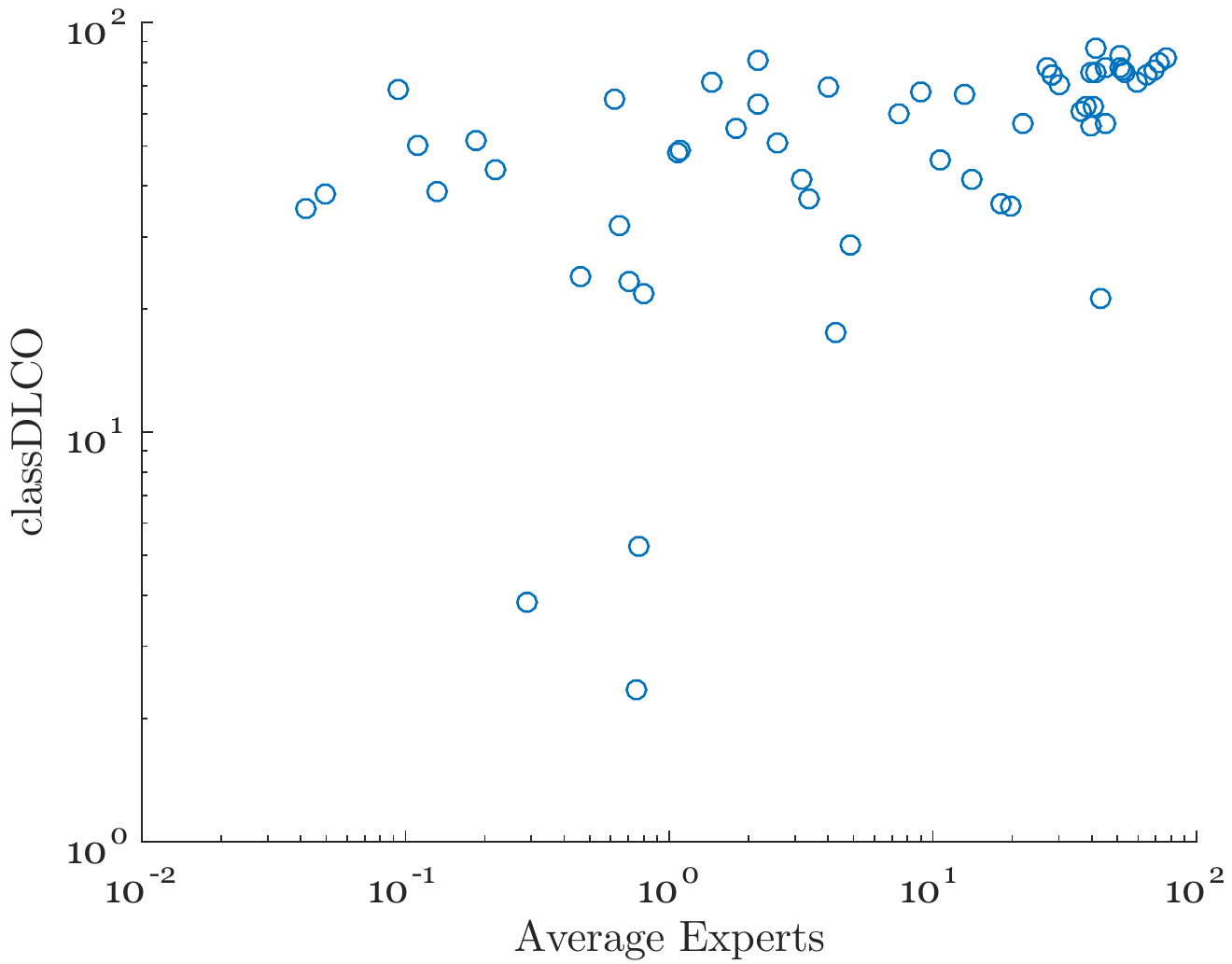}
		\includegraphics[width=0.32\textwidth]{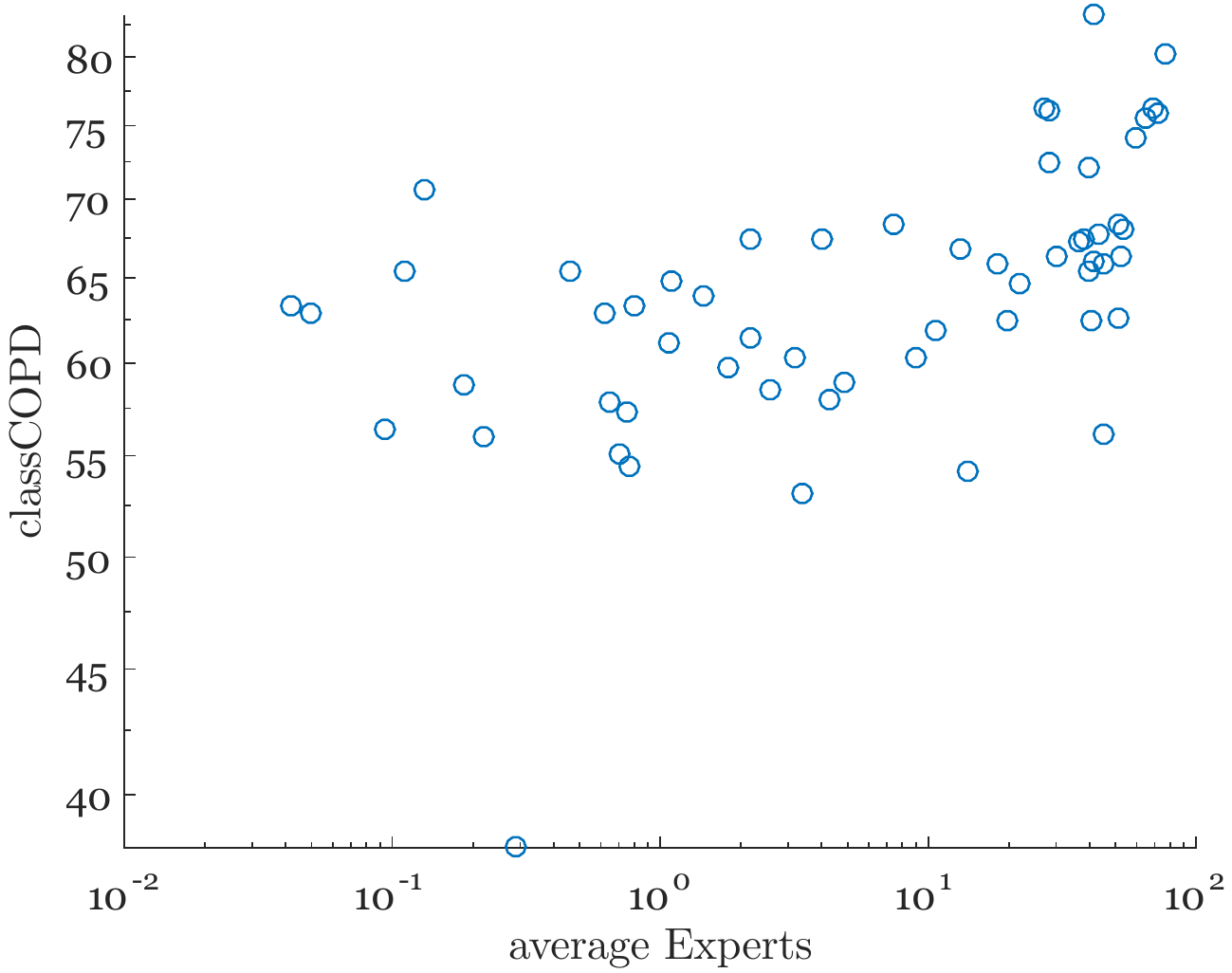}
		\includegraphics[width=0.32\textwidth]{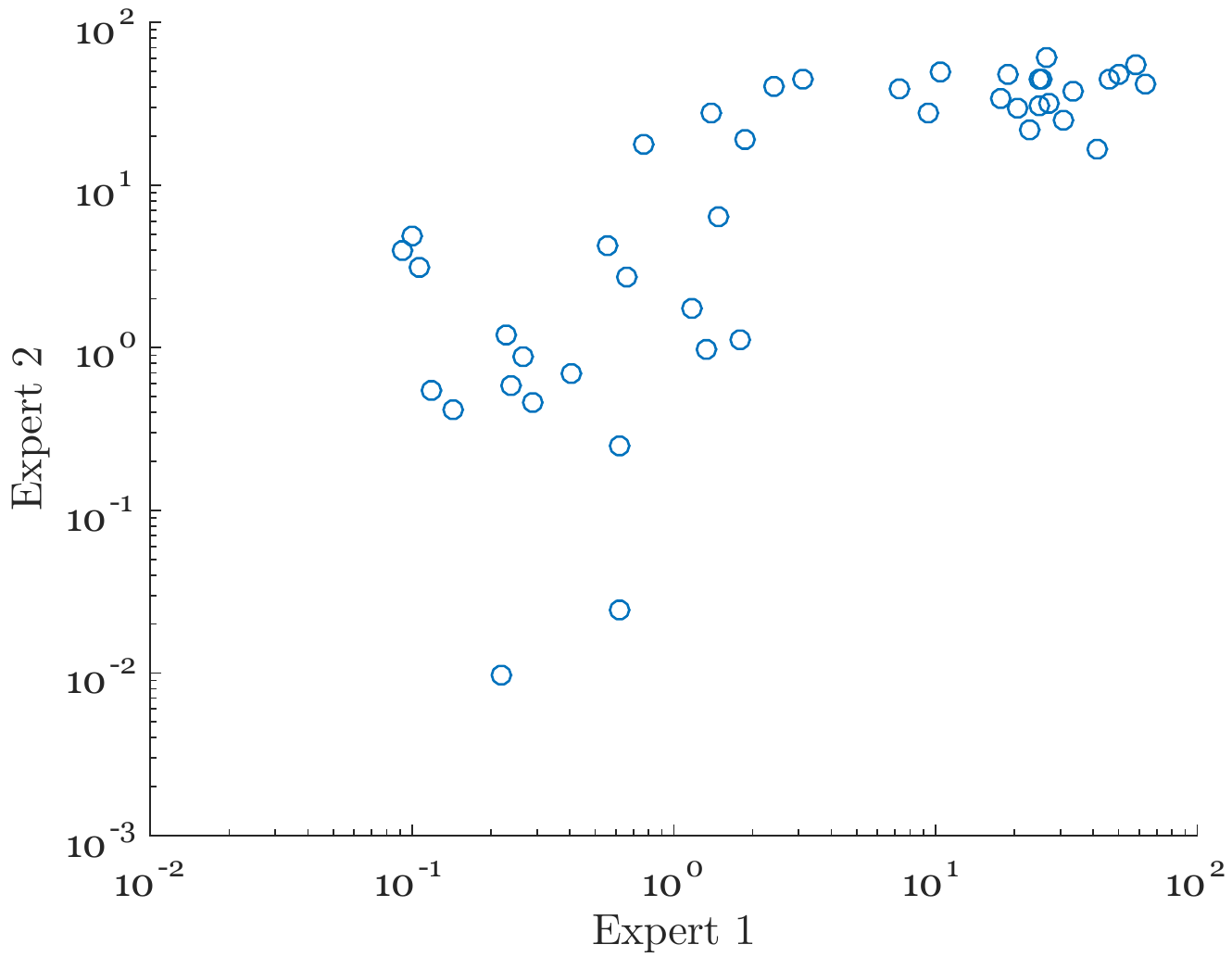}
	
	{\caption{Percentage of emphysema (log scale for visibility) per subject annotated by the experts and computed by the classifiers.}\label{fig:picScat}}
\end{figure*}

The agreements of the annotations between the two radiologists and between the classifier and the radiologists are investigated. The corresponding scatter plots between the percentage of emphysema calculated by the classifiers and the average percentage of emphysema annotated by the two radiologists are shown in Fig 3.~\ref{fig:picScat}.


Furthermore, Spearman correlation analysis are calculated between the percentage of emphysema computed from the manual annotations of the two experts (rho=0.756, $p=1.71e-14$), and the percentage of emphysema computed from the classifiers and the average percentage of emphysema from the manual annotations (ClassDLCO: rho=0.561, $p=3.01e-7$; ClassCOPD: rho=0.515, $p=4e-6$).

An example of the results from the classifiers, the manual annotations from the experts, and the threshold using LAA is presented in Fig 4. 


Although at instance level, the agreement between the classifier and the experts is not perfect, the emphysema quantification is consistent when using ClassDLCO. In contrast to ClassCOPD, ClassDLCO identifies small emphysema areas in the same patients in which the experts do not make annotations or the annotations are small, and it identifies larger emphysema areas where experts annotate large emphysema lesions.

\begin{figure*}[ht]
		\centering
		\includegraphics[width=1\textwidth]{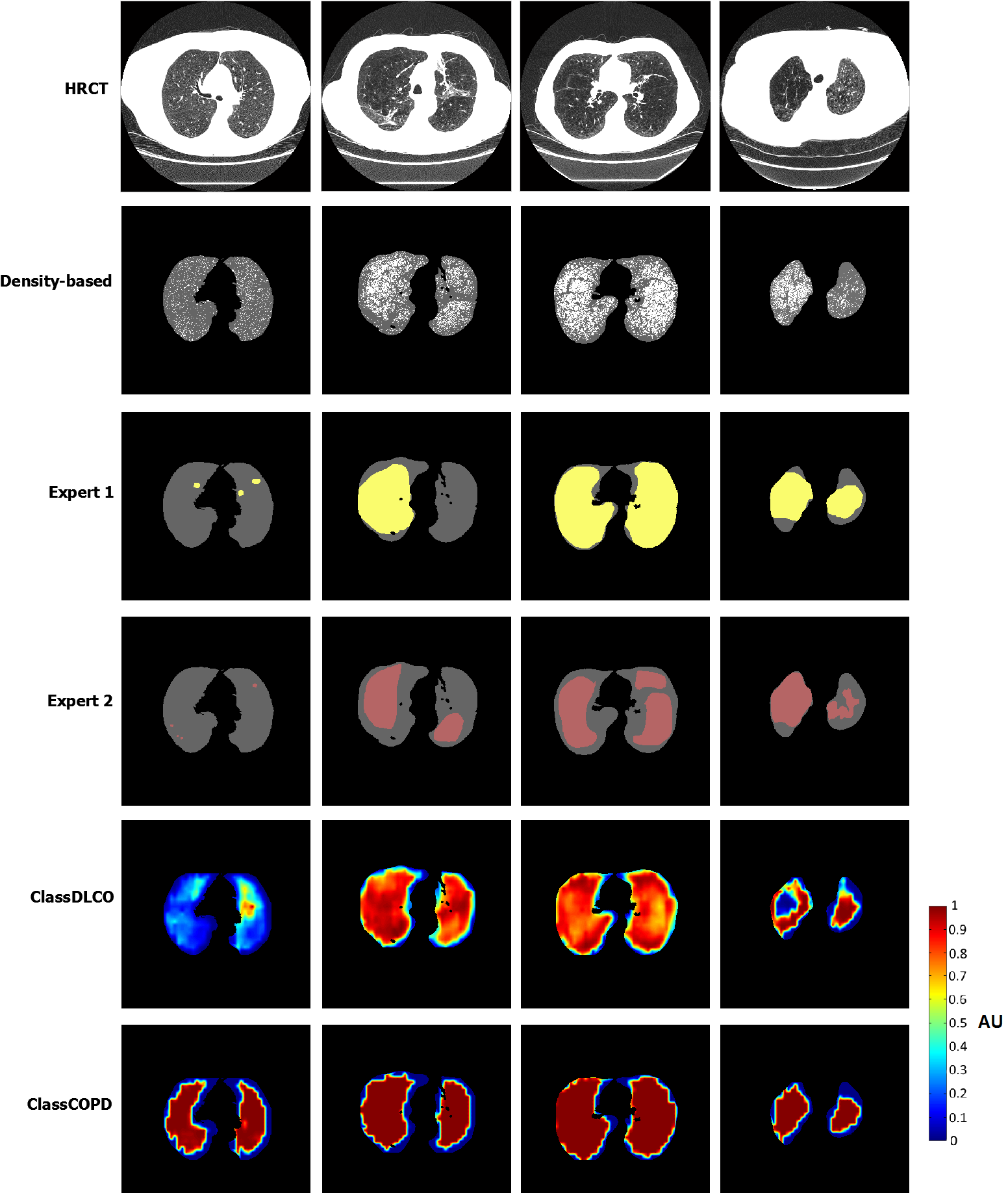}
	
	{\caption{Example of results in randomly selected slices for the density based method, manual annotations from the experts, and classifier results using miSVM-Q and Gaussian features. From left to right: patients with mild, moderate, severe and very severe COPD.}\label{fig:picClassAnn}}
\end{figure*}

The agreement between the annotations from both experts is also calculated using DSC which results in a value of 0.34, indicating a weak agreement between experts.

\section{Discussion}

\subsection{Classifier performance}
In contrast with previous studies, this study uses an MIL approach to automatically identify emphysema regions in COPD patients without requiring manually annotated HRCT scans for training. Two robust versions of the MILES and miSVM classifiers are presented. Because good bag-level (patient classification) performance does not correspond to good instance-level (patch classification) performance and vice versa~\cite{cheplygina2015label,tragante2011instance,Vanwinckelen16}, both bag-level AUC and a measure of instance-level performance, called Separability, are taken into account. The best performing classifier is miSVM-Q. Other studies~\cite{kandemir2014computer, cheplygina14} have shown that instance-level classifiers, such as miSVM, tend to have lower bag-level performance. In the present study, the bag-level performance of miSVM-Q is improved compared to the original miSVM by relaxing the condition that a bag should be classified as positive as soon as a single positive instance is detected.

The performance achieved at the bag-level is very high, with an AUC equal to 100\% for the COPD class label. This result could be explained by the fact that half of the subjects in both datasets are in the severe and very severe stage of the GOLD stratification and, therefore, these stages are easier to identify by the classifier. In ~\cite{soren12,cheplygina14} the same type of labels were used, however an AUC close to 75\% was achieved. However, the dataset used was from a screening trial, and thus contained a much higher fraction of mild COPD subjects, which were difficult to classify correctly.

\added{We used several parameters which were determined in previous studies, such as the patch size of 41x41x41 and using 50 patches to represent an entire scan. The results in the current study demonstrate that for these data, these are also reasonable choices. However, this would in general depend on a number of factors. For example, we would expect that if the cases in the dataset are all mild, a larger number of patches would be needed to capture some areas with emphysematous tissue. Another factor is the image resolution and therefore the physical size of the patch - for a smaller physical size, we expect that more patches would be needed. There are limits to this trade-off, as a much smaller physical size would not be able to capture the appearance of emphysema.}

Earlier studies~\cite{soren12,cheplygina14} did not investigate the agreement of the classifier with manual instance-level annotations. Therefore, when making choices such as patch size, number of patches per scan, and so forth, the instance-level performance was not considered. Consequently, it would be worth investigating how the patch size and number of patches (which in this study are set to the same values as in~\cite{soren12,cheplygina14}) would affect the instance-level performance.

Features derived from Gaussian filters, using both classifiers and both labels, provide larger Separability than co-occurence features or their combinations. All of these feature sets are high-dimensional compared to the size of the data. However, we observed that using lower-dimensional versions of these features reduced the performance, and high-dimensional features have been used with success in previous studies~\cite{soren12,cheplygina14}. These observations are consistent with studies of MIL classifiers on non-medical datasets, such as~\cite{cheplygina2015multiple}, where for example the Web datasets have 2K instances in 5K dimensions, but where miSVM is the second best classifier, and the best classifier that can provide instance-labels. The miSVM classifier is therefore effective at dealing with high dimensional data, by using regularization to reduce overfitting.

An interesting difference with respect to the relative dimensions is between the miSVM-Q and MILES-Q. Because miSVM-Q is an instance-level approach, its effective sample size is the total number of patches used, while the effective sample size of MILES is lower, i.e. the total number of subjects. This could explain why the performances of miSVM-Q are higher overall.
This is also consistent with previous results on similar COPD data~\cite{cheplygina14} and non-medical datasets~\cite{cheplygina2015multiple}. Therefore, MILES appears to be more prone to overfitting. It would be of interest to investigate how increasing the number of patches per scan would affect the results - for miSVM this would mean an increased sample size, but for MILES it would mean an increased dimensionality.

\added{We did not investigate the use of deep learning methods. This could be done in three ways: training a network from scratch, fine-tuning a pretrained network or using ``off-the-shelf'' features. Based on recent results in medical imaging~\cite{litjens2017survey}, we expect that training from scratch would not lead to good results due to the small dataset. We expect fine-tuning and ``off-the-shelf'' method (which could be used together with the miSVM-Q or MILES classifiers) to be more successful than training from scratch. However, both methods would depend on the data that is used to pretrain the networks, as well as several other parameters. Combining traditional features with features extracted by deep learning has also shown to be effective~\cite{nanni2017handcrafted}, and would be interesting direction for future work.}

\subsection{Clinical validation}
Spirometry has been widely used as an indicator of COPD severity due to the correlation between FEV$_1$/FVC and airway obstruction. However, FEV$_1$ does not reflect structural changes in the lung parenchyma, and therefore, it is not a reliable indicator of emphysema lesions. In contrast, DLCO is a good indicator of the level of anatomic emphysema. In this study, a Spearman correlation analysis between the best classifier from the classifier evaluation, miSVM-Q, and the PFTs is computed. The results in the present study show, as presented in Table \ref{tab:SpearmanClass}, that the classifier using both labels has a higher correlation with DLCO values than with FEV$_1$. This result is comparable to the result in~\cite{Tan11}, where the emphysema segmentation using a texture-based approach had a better correlation with DLCO than with values from FEV$_1$. This is explained because FEV$_1$ measures airflow obstruction; however, this is only partially reflected in emphysema lesions. The classifier that is trained on the COPD label based on FEV$_1$ values, likely detects mostly signs of emphysema and therefore, still correlates better with DLCO than with FEV$_1$.


The correlations from the classifier with the PFTs are also compared with the correlations between the PFTs and a density mask method that has been widely used to quantify emphysema lesions in CT scans. The results show that the density based method correlates moderately better with FEV$_1$ than both the classifiers and the expert evaluations, and the same behaviour can be observed in ~\cite{Tan11}. This may be explained by the inability of the density mask to discriminate between air trapping and emphysema due to the nature of their thresholds~\cite{Voelkel08}. However, other studies from the literature that aim to quantify emphysema~\cite{park08, soren12} show a better correlation between FEV$_1$ with their proposed texture analysis methods than a traditional density based method.

This study uses the PFTs as the most reliable measurement to validate the results of the classifier despite manual annotations by two independent experts being available. This is due to the weak agreement between experts in the annotations of emphysema, as shown by the Spearman correlation results in Section~\ref{ManAnn} and the Dice similarity coefficient. This is in agreement with~\cite{Mascalchi12}, who showed a low inter-observer agreement in a task of quantifying emphysema in whole lungs. As shown in Fig 3,
ClassCOPD overestimates the amount of emphysema in comparison with the manual annotations. This result can be produced by the small dataset of non-COPD. However, ClassDLCO generally tends to agree with experts on the size of emphysema areas. The scatter plots show that ClassDLCO has a good agreement with the experts' annotations in severe and very severe cases, but the agreement is fair in moderate patients and poor in mild patients. ClassDLCO overestimates the emphysema in these cases.

However, these findings in conjunction with the improved correlation with DLCO compared to manual annotations could indicate that ClassDLCO is more sensitive to early changes in the lung parenchyma and can detect emphysema even before these changes are able to be detected visually. To confirm this result, future studies should investigate the progression of emphysema in the areas where the classifier finds emphysema, but that were not assessed visually. This will also help to reduce inter-observer variability, which is a major limitation in visual assessment, as other studies have reported~\cite{Ginsburg12, Barr12}. Furthermore, the correlation results between the ClassDLCO and DLCO values show that quantitative assessment of emphysema with the presented method provides an important measurement of the reduction in the alveolar area. In addition, as suggested in~\cite{Diaz15}, a better detection of emphysema in HRCT scans can also be used in refining the prediction of the 6 minute walking distance test.

\subsection{Limitations}
A limitation of this study is the size and balance of the datasets. The Fre dataset is very small, and the Aal dataset does not contain any controls. Due to this imbalance, the DLCO threshold used is lowered to 60\%; thus, patients with lower DLCO values are included in the high DLCO class (which could be seen as healthy, although note that the COPD diagnosis is based on spirometry). The texture features extracted from the non-COPD group could have a similar representation as the features extracted from COPD patients because different lung diseases could appear in CT scans as LAA as emphysema does, i.e. cystic lung disease. Therefore, it would be desirable to include scans without any pathology.

A related problem is the fact that the Fre and Aal datasets have different acquisition parameters, which can negatively affect classification performance~\cite{de2016machine}. An improvement in performance will be expected if the appearance of healthy tissue could be learned from both datasets rather than only Fre. An alternative would be to use techniques such as intensity normalization or transfer learning classifiers to reduce the differences between datasets.

\section{Conclusion}

This study presented two new versions of multiple instance classifiers which identify emphysema regions in patients suffering from COPD without requiring manual annotations. On a clinical dataset with 88 subjects in total, the proposed method showed a good correlation with the pulmonary function tests, particularly with DLCO. 
The proposed method had a moderate correlation with manual annotations of emphysema, however, this correlation was higher than that of a density based method, which was also moderately correlated with manual annotations. 
Therefore it could be considered as a reliable tool to support radiologists in the assessment of emphysema to reduce the inter- and intra-observer variability. 
As future work, validating the results on a larger and more balanced dataset, and investigating the effect of different acquisition parameters, are recommended.

\section*{Disclosure of Conflicts of Interest}
The authors have no relevant conflicts of interest to disclose. We gratefully acknowledge financial support from the Netherlands Organization for Scientific Research (NWO), grant no. 639.022.010.

\bibliographystyle{plos2015}
\bibliography{refs}

\end{document}